\documentclass{article}

\PassOptionsToPackage{numbers, compress}{natbib}

\usepackage[preprint]{neurips_2025}

\usepackage[utf8]{inputenc} 
\usepackage[T1]{fontenc}    
\usepackage{hyperref}       
\usepackage{url}            
\usepackage{booktabs}       
\usepackage{amsfonts}       
\usepackage{amsmath}        
\usepackage{nicefrac}       
\usepackage{microtype}      
\usepackage{xcolor}         
\usepackage{xpatch}
\usepackage{graphicx}
\usepackage{subcaption}
\usepackage{arydshln}  
\usepackage{tikz}      
\usepackage{adjustbox} 
\usepackage{amssymb}
\usepackage{amsthm}
\usepackage{algorithm}
\usepackage{algorithmic}
\usepackage{bbding}

\usepackage{pifont}
\usepackage{wrapfig}
\usepackage{colortbl}
\usepackage{multirow}
\usepackage{xpatch}
\makeatletter
\xapptocmd{\NAT@bibsetnum}{\setlength{\leftmargin}{0pt}\setlength{\itemindent}{\labelwidth}\addtolength{\itemindent}{\labelsep}}{}{}
\makeatother

\title{Uni4D: A Unified Self-Supervised Learning Framework for Point Cloud Videos}

\author{
Zhi Zuo$^{1*}$, Chenyi Zhuang$^{2*}$, Pan Gao$^{1\dagger}$, Jie Qin$^{1\dagger}$, Hao Feng$^{3}$, Nicu Sebe$^{2}$ \\
$^1$ Nanjing University of Aeronautics and Astronautics \\
$^2$ University of Trento, 
$^3$ Central China Normal University \\
}

\begin{document}
\renewcommand{\thefootnote}{\fnsymbol{footnote}}
\footnotetext{$^*$Equal contribution. 
$^\dagger$Corresponding author. }
\renewcommand{\thefootnote}{\arabic{footnote}}

\maketitle

\begin{abstract}
Self-supervised representation learning for point cloud videos remains a challenging problem with two key limitations: (1) existing methods rely on explicit knowledge to learn motion, resulting in suboptimal representations; (2) prior Masked AutoEncoder (MAE) frameworks struggle to bridge the gap between low-level geometry and high-level dynamics in 4D data. In this work, we propose a novel self-disentangled MAE for learning expressive, discriminative, and transferable 4D representations. To overcome the first limitation, we learn motion by aligning high-level semantics in the latent space \textit{without any explicit knowledge}. To tackle the second, 
we introduce a \textit{self-disentangled learning} strategy that incorporates the latent token with the geometry token within a shared decoder, effectively disentangling low-level geometry and high-level semantics. In addition to the reconstruction objective, we employ three alignment objectives to enhance temporal understanding, including frame-level motion and video-level global information. We show that our pre-trained encoder surprisingly discriminates spatio-temporal representation without further fine-tuning. Extensive experiments on MSR-Action3D, NTU-RGBD, HOI4D, NvGesture, and SHREC'17 demonstrate the superiority of our approach in both coarse-grained and fine-grained 4D downstream tasks. Notably, Uni4D improves action segmentation accuracy on HOI4D by $+3.8\%$. 

\end{abstract}

\section{Introduction}
\label{sec:intro}
Self-Supervised Learning (SSL) has emerged as a powerful approach for vision understanding, enabling high-quality pre-trained models with strong downstream performance and adaptability after fine-tuning~\cite{he2020momentum, he2022masked, chen2020simple, kenton2019bert,tong2022videomae, feichtenhofer2022masked, PointContrast2020, pang2022masked}. This success has recently extended to point cloud video understanding, which requires modeling the spatial and temporal dynamics of 4D data~\cite{sheng2023contrastive,shen2023pointcmp,sheng2023point,zhang2023complete,shen2023masked}. Typically, spatial representation encodes the \textit{geometry} or \textit{appearance} of a 3D scene, capturing structural and visual details of 3D objects at a specific moment. In contrast, temporal representation focuses on the \textit{motion} or \textit{dynamic changes} across consecutive frames, such as object movements or transformations. However, learning expressive 4D representations from point cloud videos remains highly challenging due to their unstructured and dynamic nature. Prior studies resort to masked reconstruction to jointly learn spatio-temporal features, but still fail to capture long-range temporal dependencies, which are essential for fine-grained 4D understanding tasks~\cite{Liu_2022_CVPR}. 

\begin{wrapfigure}{r}{0.5\textwidth}
\vspace{-4mm}
\begin{center}
\includegraphics[clip,trim=0 0 0 0,width=0.46\textwidth]{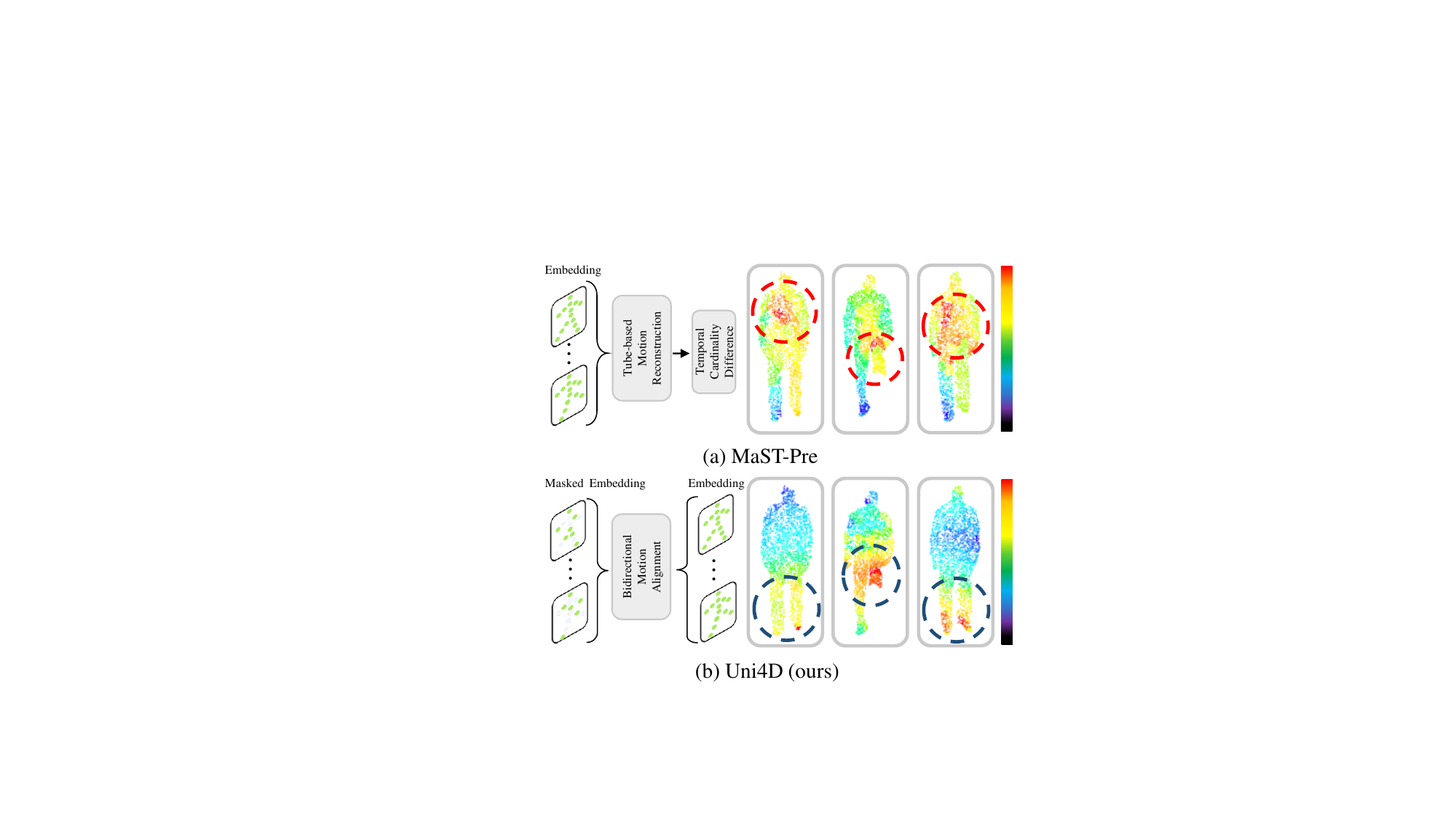}
\end{center}
\vspace{-2mm}
\caption{Comparison of motion patterns learned by the pre-trained encoder (without fine-tuning). Uni4D shows consistent activation in the foot region during ``kick forward", indicating a capture of long-term temporal dependencies.}
\vspace{-2mm}
\label{fig:motivation} 
\end{wrapfigure}

To extract motion from point cloud videos, existing MAE-based methods often rely on \textit{explicit knowledge}. MaST-Pre \cite{shen2023masked} introduces spatio-temporal prediction and reconstructs masked point clouds. This method effectively captures appearance information, while hand-crafting temporal cardinality difference to represent motion, which partitions points into spherical octants and computes their angular distances to the central line. To further investigate its impact, we extract the attention activation from the pre-trained encoder on three consecutive frames and visualize in Fig. \ref{fig:motivation}. Obviously, this design leads to inconsistent motion patterns, indicating a limited ability to model long-term temporal dependencies. In Tab.~\ref{tab:motion_compare}, MaST-Pre exhibits a significant performance drop in linear probing, where the encoder is frozen and only a classifier is trained. This observation verifies that the pre-trained encoder of MaST-Pre fails to capture meaningful 4D representations. Similarly, M2PSC~\cite{han2024masked} leverages an external network to guide motion trajectory prediction. Although both methods show effectiveness, such explicit knowledge of motion undermines the learning ability of the encoder.

\begin{table}[t]
\centering
\caption{Comparison of motion modeling strategies of recent MAE-based point cloud video SSL methods. We report pre-training efficiency and fine-tuning performance on MSR-Action3D.}
\vspace{1mm}
\resizebox{\textwidth}{!}{
\begin{tabular}{lccccccccc}
\toprule[1pt]
\multirow{2}{*}{Method} & \multicolumn{3}{c}{Motion Modeling Design} & \multicolumn{4}{c}{Pre-train Efficiency} & \multicolumn{2}{c}{Performance} \\
\cmidrule(lr){2-4} \cmidrule(lr){5-8} \cmidrule(lr){9-10}
& Strategy & Space & Explicit Knowledge & \# Params & GFLOPS & epoch/s & Branch & Fine-tuning & Linear Probing \\
\midrule
Point-MAE~\cite{pang2022masked} & Reconstruction & Euclidean & - & 55.8M & 25.3 & 17 & One & 85.36 & 71.08 \\
MaST-Pre~\cite{shen2023masked} & Reconstruction & Euclidean & Hand-crafted & 55.3M & 10.4 & 17 & One & 91.29 & 63.80 \\
M2PSC~\cite{han2024masked} & Reconstruction & Euclidean & External priors & 94.5M & 57.5 & - & One & 93.03 & - \\
Uni4D & Alignment & Latent & - & 100.4M & 34.3 & 26 & Two & \textbf{93.38} & \textbf{84.62} \\
\bottomrule[1pt]
\end{tabular}
}
\vspace{-4mm}
\label{tab:motion_compare}
\end{table}

This motivates our central question: \textbf{Can motion be learned in a fully self-supervised manner, without any explicit knowledge?} Our key insight is that the latent space offers an ideal setting for capturing high-level semantics such as motion \cite{preechakul2022diffusion, dong2022bootstrapped, jeong2024training}. One promising direction is to adapt MAE-based frameworks~\cite{huang2023contrastive,chen2022sdae,chen2024context}, separating motion alignment in the latent space and geometry reconstruction in Euclidean space.
However, in practice, the decoder tends to overfit to low-level geometric details due to the inherent complexity of 4D data. The gap between geometry and motion makes it particularly challenging for the decoder to perform low-level reconstruction while preserving high-level semantics without any explicit motion cues. Overcoming the dilemma between semantic modeling and masked reconstruction is crucial for the encoder to learn rich 4D representations. Nevertheless, this challenge remains largely underexplored in point cloud video understanding.
 
To address these, we propose a novel approach to effectively \textit{capture}, \textit{disentangle}, and \textit{discriminate} 4D representations in a self-supervised manner, termed \textbf{\textit{self-disentangled MAE}}. Our approach specifically tackles two major limitations in prior works: (1) To model motion without relying on explicit knowledge, we align high-level semantics in the latent space, including a bidirectional alignment objective for learning motion and a global alignment objective for capturing video-level semantics. (2) To avoid overfitting to geometric details and preserve semantics during decoding, we propose a novel self-disentangled learning strategy, with two learnable tokens appended into a shared decoder to disentangle low-level and high-level features with two independent objectives. Surprisingly, our encoder already models highly discriminative 4D representations through pre-training alone. It demonstrates remarkable fine-tuning performance improvements across a variety of downstream 4D tasks, yielding the name \textbf{Uni4D}. The main contributions of this paper are: (1) A novel self-supervised learning framework for point cloud video understanding that learns high-level semantics in the latent space ; (2) An efficient strategy using two learnable tokens to disentangle high-level and low-level features; (3) Extensive experiments validate the superiority of our approach with notable improvements on diverse 4D benchmarks, ranging from coarse-grained recognition to fine-grained segmentation. 

\section{Related Works}
\label{related works}
\textbf{Visual Mask Reconstruction}. Masked modeling has greatly succeeded in natural language processing~\cite{kenton2019bert,brown2020language}. Numerous computer vision tasks~\cite{bao2021beit,wang2022bevt, he2022masked} adopted MAE-based frameworks and significantly improved performance in downstream tasks, suggesting the effectiveness of pre-training~\cite{chen2022sdae, dong2022bootstrapped,chen2024context}. This paradigm is subsequently adapted to video understanding ~\cite{tong2022videomae,feichtenhofer2022masked, wei2022masked, chen2022sdae} and point cloud representation learning~\cite{yu2021pointbert,pang2022masked,zhang2022point,liu2022masked,yan2024d}. The representation of dynamic point clouds is essential for countless applications in computer graphics. Unlike voxel-based~\cite{luo2018fast,choy20194d,wang20203dv} and point-based~\cite{liu2019meteornet,fan2021pstnet,fan2021deep,fan2019pointrnn,min_CVPR2020_PointLSTM,fan2021point,fan2022point,wen2022point, Zhong_2022_CVPR, Ben-Shabat_2024_CVPR} representation learning methods, point cloud SSL approaches do not rely heavily on the quality and scale of labeled datasets and ease the expensive data annotations. Other learning techniques, including contrastive learning \cite{radford2021learning, PointContrast2020,afham2022crosspoint}, have also been widely studied.

\textbf{Self-supervised Learning on Point Cloud Videos}. Early efforts in self-supervised learning for point cloud videos primarily adopt contrastive learning frameworks~\cite{sheng2023contrastive, shen2023pointcmp, sheng2023point, zhang2023complete}. More recently, masked modeling approaches such as MaST-Pre~\cite{shen2023masked} and M2PSC~\cite{han2024masked}, have emerged, which reconstruct masked point cloud tubes to learn spatio-temporal representations. To model motion, MaST-Pre introduces a temporal cardinality difference task, while M2PSC predicts motion trajectories using an auxiliary pre-trained network. However, both methods involves computationally expensive pre-processing, which is less generalizable. More importantly, their motion modeling relies on explicit knowledge, resulting in low-quality 4D representations that are insufficient for fine-grained 4D downstream tasks. In contrast, our work innovatively align high-level semantics in the latent space, enabling the learning of long-term temporal representations in a fully self-supervised manner, without any hand-crafted feature extraction or external models.

\section{Method}
\label{sec:Methods}
\begin{figure*}[t]
    \centering
    {\includegraphics[width=1\textwidth]{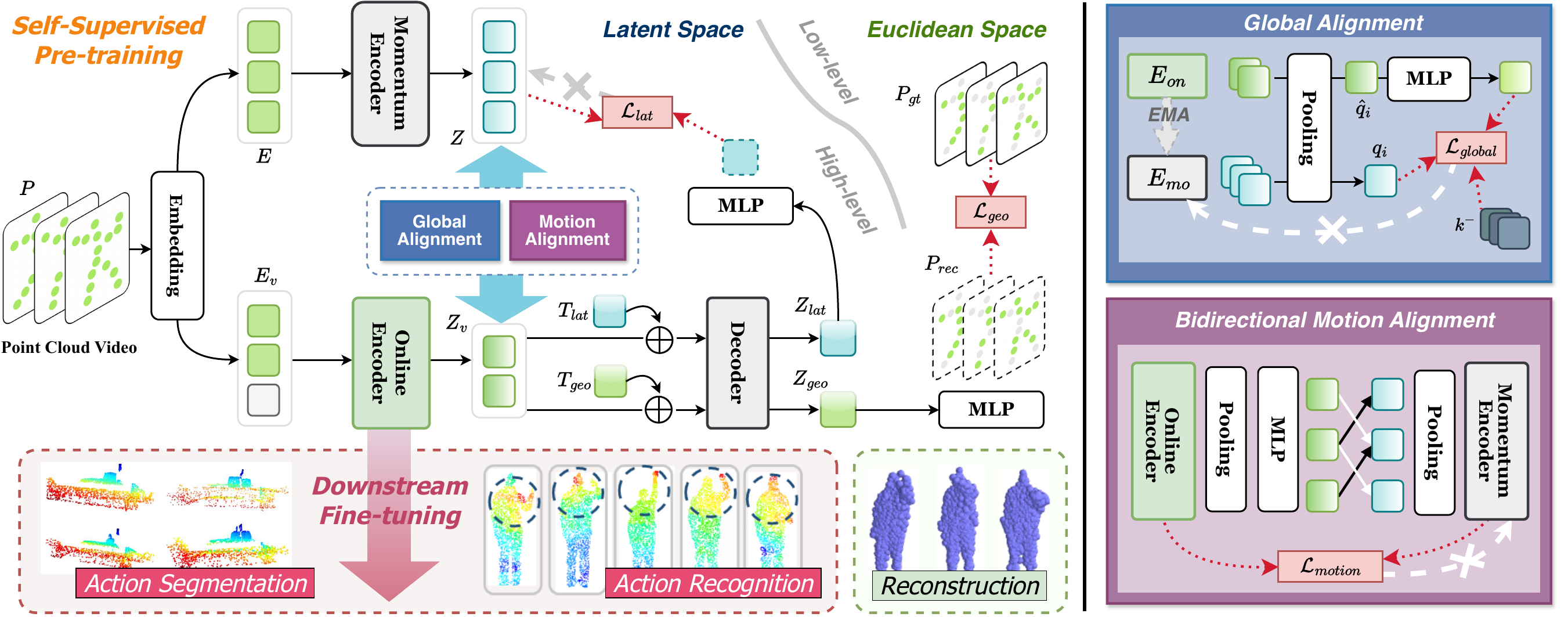}}
	\caption{Overview of the Uni4D framework. Our approach aligns the motion and global semantics in the latent space while reconstructing geometry in Euclidean space. Two learnable tokens are used to disentangle low-level and high-level features during decoding. The pre-trained encoder learns discriminative 4D representations to boost various 4D downstream tasks.}
	\label{fig:pipeline} 
    \vspace{-3mm}
\end{figure*}

Our ultimate goal is to pre-train a high-quality encoder that learns meaningful 4D representations for fine-tuning on broader tasks. As illustrated in Fig. \ref{fig:pipeline}, our approach consists of an online encoder along with a decoder to reconstruct masked point tubes, and a momentum encoder to provide additional prediction targets. Two high-level semantics, frame-level motion and video-level global information, are aligned in the latent space while geometry is learned in Euclidean space. The self-disentangled learning ability is achieved by two learnable tokens appended into the decoder, which prevent feature entanglement and semantic diminishment. Without any explicit knowledge, Uni4D self-supervises the learning of discriminative 4D representations during pre-training.

\subsection{Masked Modeling}
Consider the input of a point cloud video as $\boldsymbol{P}\in \mathbb{R}^{L\times N\times 3}$, where $L$ is the sequence length and $N$ denotes the number of points in each frame. Following \cite{shen2023masked}, we construct the point tube sequence from the input point cloud sequence as:
\begin{equation}
    \boldsymbol{Tube}_{\hat{p}_i} = \{ p| p \in \boldsymbol{P}, S(p,\hat{p}_i)< r_s, T(p,\hat{p}_i)\leq \frac{r_t}{2}\},
\end{equation} 
where $\hat{p}_i$ is sampled from $p$ using Farthest Point Sampling (FPS), $S$ is the Euclidean distance, $T$ is the difference of two points in frame timestamps, $r_s$ and $r_t$ represent the spatial radius and the frame length in a tube, respectively. Next, these point tubes are projected to the full tube embeddings by Point 4D convolution (P4D)~\cite{fan2021point}, denoted as $\boldsymbol{E}$. To perform masked reconstruction, we randomly mask the point tubes using a high ratio (default to 75\%) to ease the learning of expressive representations and benefit the modeling of a meaningful latent space. The above process yields $N_v$ visible embeddings, denoted as $\boldsymbol{E}_{v}$. Appendix~\ref{theory} presents the embedding process in more detail.

\subsection{Architecture Design}
We employ an autoencoder for masked modeling, along with a momentum encoder to guide semantic modeling. Below, we present the architectural details of the proposed Uni4D framework.

\textbf{Online Encoder}. The online encoder is implemented as a spatial-temporal Transformer~\cite{fan21p4transformer}. This encoder only handles visible embeddings $\boldsymbol{E}_{v}$ and encodes them into embeddings $\boldsymbol{Z}_{v}\in\mathbb{R}^{L'\times N_v\times C}$ in the latent space, where $L'$ is the output length of P4D. 

\textbf{Momentum Encoder}. To learn the latent representation in a self-supervised manner, we are inspired by \cite{dong2022bootstrapped, chen2022sdae} to adopt a momentum encoder and update its weights from the online encoder using Exponential Moving Average (EMA). We input the unmasked full embeddings into the momentum encoder, i.e., $z^0=\boldsymbol{E}$, to ensure intact representation extraction. We denote the outputs of the momentum encoder as $\boldsymbol{Z}\in\mathbb{R}^{L'\times N'\times C}$. We provide more details of architecture in Appendix~\ref{appendix:architecture}.

\textbf{Decoder}. We employ a lightweight decoder $\mathcal{D}$ after the online encoder for masked modeling. To reconstruct masked embeddings, several geometry tokens are added to the decoder, denoted $\boldsymbol{T}_{geo}$. Our self-disentangled learning is achieved by appending the latent tokens, decoded $\boldsymbol{T}_{lat}$, to preserve high-level features during decoding. Note that $\boldsymbol{T}_{geo}$ and $\boldsymbol{T}_{lat}$ are both learnable and concatenated to the visible embeddings $\boldsymbol{E}_v$ within a shared decoder. However, they function differently during decoding: $\boldsymbol{T}_{geo}$ is to represent the masked embeddings for low-level reconstruction in Euclidean space, while $\boldsymbol{T}_{lat}$ is to preserve high-level features for alignment in the latent space.

\subsection{Self-Disentangled Learning Strategy}
\label{sec:self-disentangled strategy}
As discussed in Sec.~\ref{sec:intro}, the decoder has raised the overfitting issue due to the representation gap. To address this, we introduce a self-disentangled learning strategy, which leverages two kinds of learnable tokens: the geometry tokens to prompt the decoder for low-level geometric reconstruction, and latent tokens t to prompt the decoder for preserving high-level semantics.

\subsubsection{Low-level Reconstruction Objective}
To conduct masked reconstruction, we append the geometry tokens $\boldsymbol{T}_{geo}$ to the encoded visible embeddings $\boldsymbol{Z}_v$ into the decoder and obtain $\boldsymbol{Z}_{geo}=\mathcal{D}(\boldsymbol{T}_{geo};\boldsymbol{Z}_v)$. Next, we apply a 1D convolution as the prediction head $\mathcal{H}$ to obtain final predictions, denoted $\boldsymbol{P}_{rec}=\mathcal{H}(\boldsymbol{Z}_{geo})$. We adopt the $l_2$ Chamfer Distance loss (CD) between $\boldsymbol{P}_{rec}$ and the ground truth $\boldsymbol{P}_{gt}$:
\begin{equation}
\label{eq:geometry_loss}
\begin{aligned}
\mathcal{L}_{geo} = \frac{1}{r_t}\sum^{r_t}_{i=1}\Bigg\{\frac{1}{\left|\boldsymbol{P}_{rec}^i \right|} \sum_{x \in \boldsymbol{P}_{rec}^i} \min_{y \in \boldsymbol{P}_{gt}^i} \|x - y\|^2 + \frac{1}{\left|\boldsymbol{P}_{gt}^i \right|}\sum_{y \in \boldsymbol{P}_{gt}^i } \min_{x \in \boldsymbol{P}_{rec}^i } \|y - x\|^2\Bigg\},
\end{aligned}
\end{equation}
where $r_t$ is the frame length in each tube.

\subsubsection{High-level Alignment Objective}
To prevent overfitting to low-level geometry during decoding, we introduce a latent alignment objective in addition to the reconstruction objective. Next, two independent alignment objectives in the latent space aim to extract the frame-level consistent motion and video-level global information.

\textbf{Latent Alignment}. We aim to encourage the decoder to preserve high-level semantics while reconstructing geometric structures. Without introducing another decoder, we append the latent tokens $\boldsymbol{T}_{lat}$ into the same decoder to effectively disentangle high-level semantics from low-level information. Simply, we denote the output of the decoder as $\boldsymbol{Z}_{lat}=\mathcal{D}(\boldsymbol{Z}_v,\boldsymbol{T}_{lat})$. To supervise the decoder for latent alignment, we employ the following objective:
\begin{equation}
\label{eq:latent_loss}
    \mathcal{L}_{lat} = 1-cos(MLP(\boldsymbol{Z}_{lat}),\texttt{stopgrad}(\boldsymbol{Z})),
\end{equation}
where $cos(\cdot)$ denotes the cosine similarity, $\boldsymbol{Z}$ is the output of the momentum encoder. Notice that the projection is applied only to $\boldsymbol{Z}_{lat}$. Therefore, the latent token $\boldsymbol{T}_{lat}$ is effectively optimized to prompt the decoder to preserve high-level features aligned with the latent embeddings. Stop-gradient operation \cite{chen2021exploring} is employed to prevent the momentum encoder from collapsing.

\textbf{Bidirectional Motion Alignment}. Unlike \cite{shen2023masked, sheng2023contrastive}, we aim to capture temporal dependencies directly in the latent space without explicit knowledge. Specifically, two latent embeddings $\boldsymbol{Z}_v$ from the online encoder and $\boldsymbol{Z}$ from the momentum encoder are reshaped and applied max-pooling operation over the spatial dimension, yielding frame-level features $\boldsymbol{H}_v=\{\hat{h}_1,\hat{h}_2,...,\hat{h}_{N'}\}$ and $\boldsymbol{H}=\{h_1,h_2,...,h_{N'}\}$, with shape $\mathbb{R}^{L'\times 1\times C}$. Inspired by \cite{zhang2023complete}, we employ a bidirectional motion alignment objective in the latent space to model consistent motion patterns, which is formulated as:
\begin{equation}
\label{eq:motion_loss}
    \mathcal{L}_{motion} = - \frac{1}{B} \sum_{i=1}^{B} \log \frac{\exp\left(  MLP(\hat{h}_i) \cdot h_{j} / \tau \right)}{\sum_{k\neq j} \exp\left( MLP(\hat{h}_i) \cdot h_{k} / \tau \right)},
\end{equation}
where two MLP blocks are trained to project each frame in $\boldsymbol{H}_v$ to align with its target in $\boldsymbol{H}$ from forward and backward directions, respectively. We designate $i\in[2,N']$ and $j=i-1$ for the forward alignment, while $i\in[1,N'-1]$ and $j=i+1$ for the backward alignment. Note that the momentum encoder is not updated via gradient backpropagation from this objective.

\textbf{Global Alignment}. Empirically, we find that modeling only the motion leads to inferior fine-tuning performance on coarse-grained tasks (see Fig.~\ref{fig:CELOSS}). We are driven to extract other high-level semantics in the latent space, i.e., the video-level global information. To achieve this, we reshape $\boldsymbol{Z}_v$ to $\mathbb{R}^{L'\cdot N_v\times C}$ and $\boldsymbol{Z}$ to $\mathbb{R}^{L'\cdot N'\times C}$. Next, we apply the max-pooling operation to obtain $\hat{q}\in\mathbb{R}^{1\times C}$ and $q\in\mathbb{R}^{1\times C}$. To encourage the modeling of global representations, we align two embeddings by:
\begin{equation}
\label{eq:global_loss}
    \mathcal{L}_{global} = - \log \frac{\exp (MLP(\hat{q})\cdot q / \tau) }{ \sum_{k=0}^{K} \exp\left( MLP(\hat{q})\cdot q_k / \tau \right)},
\end{equation}
where $q_k$ denotes negative samples in a queue, $K$ is the queue length. We also apply the stop-gradient operation to the momentum encoder for this objective. 

\subsubsection{Overall Loss}
Finally, we combine all objectives and define the overall loss function as:
\begin{equation}
    \mathcal{L}_{total} = \mathcal{L}_{geo} + \mathcal{L}_{lat} + \mathcal{L}_{global} + \mathcal{L}_{motion}.
\end{equation}

Without any explicit motion cues, our encoder must \textit{capture} meaningful spatio-temporal features from point cloud videos. To empower the decoder to \textit{disentangle} features for both semantic and masked modeling, the encoder needs to \textit{discriminate} high-level semantics from low-level information. As a result, our pre-trained encoder can learn comprehensive 4D representations, ultimately leading to effective downstream fine-tuning, which we demonstrate in the next section.

\section{Experiments}
\label{exp}
We compare Uni4D to state-of-the-art methods through extensive experiments, including end-to-end fine-tuning, semi-supervised learning, transfer learning, and few-shot learning. The evaluation is conducted on five benchmarks, MSR-Action3D~\cite{2010Action}, NTU-RGBD~\cite{shahroudy2016ntu}, HOI4D~\cite{Liu_2022_CVPR}, NvGestures~\cite{molchanov2016online}, and SHREC'17~\cite{de2017shrec}, covering action recognition, gesture recognition, and action segmentation. 

\subsection{Pre-training Setups}
\label{pre}
For point cloud videos in MSR-Action3D and NTU-RGBD, 24 frames are densely sampled and 1024 points are sampled from each frame. The frame sampling stride is set to 1/2 augmented with random scaling. We pre-train Uni4D on MSR-Action3D/NTU-RGBD for 200/100 epochs, respectively. For videos in HOI4D, we pre-train Uni4D for 50 epochs, with 150 frames densely sampled and 2048 points selected in each frame. We adopt P4Transformer~\cite{fan2021point} with 5/10 layers as our encoder on MSR-Action3D/NTU-RGBD. We use PPTr~\cite{wen2022point} as our encoder for action segmentation on HOI4D. The decoder is implemented as a 4-layer Transformer for all datasets. The temperature $\tau$ is set to $0.1$ for Eq.~\ref{eq:motion_loss} and Eq.~\ref{eq:global_loss}. All experiments are conducted on two GeForce RTX 4090. Appendix~\ref{appendix:details} provides more details on experimental setups for both pre-training and fine-tuning.

\subsection{End-to-end Fine-tuning}
\label{fine-tuning}
To conduct end-to-end fine-tuning, we discard the part of the pre-trained network after the encoder and replace it with a task-specific classifier. The pre-trained encoder is fine-tuned on the NTU-RGBD, MSRAction-3D, and HOI4D datasets, respectively. Without special instructions, we use the same dataset for pre-training and fine-tuning.

\begin{wraptable}{r}{7cm}
\vspace{-0.2cm}
\centering
\footnotesize
\caption{Action recognition accuracy ($\%$) on MSR-Action3D. $\dagger$ indicates pre-training on NTU.}
\label{tab:msr}
  \begin{tabular}{lc}
    \toprule
    Method  & Accuracy \\
    \midrule
    \multicolumn{2}{c}{\textit{Supervised Learning Only}} \\
    \midrule
    MeteorNet~\cite{liu2019meteornet} & 88.50 \\
    kinet~\cite{Zhong_2022_CVPR}  & 93.27 \\
    PSTNet~\cite{fan2021pstnet} & 91.20 \\
    PPTr~\cite{wen2022point}  & 92.33 \\
    3DInAction~\cite{Ben-Shabat_2024_CVPR}  & 92.23 \\
    MAMBA4D~\cite{liu2024mamba4defficientlongsequencepoint} & 92.68 \\
    \rowcolor{gray!10}
    P4Transformer~\cite{fan2021point} & 90.94 \\
    \midrule
    \multicolumn{2}{c}{\textit{with Self-Supervised Representation Learning}} \\
    \midrule
    PSTNet + PointCPSC~\cite{sheng2023point} &  92.68 \\
    PSTNet + CPR~\cite{sheng2023contrastive} &  93.03 \\
    PSTNet + PointCMP~\cite{shen2023pointcmp} &  93.27 \\
    \rowcolor{gray!10}
    P4Transformer + MaST-Pre~\cite{shen2023masked} &  91.29 \\
    \rowcolor{gray!10}
    P4Transformer + M2PSC~\cite{han2024masked} &  93.03 \\
    \rowcolor{gray!30}
    \textbf{P4Transformer + Uni4D}  & \textbf{93.38} 
    \scriptsize (\textcolor{red}{+2.44}) \\  
    \rowcolor{gray!30}
    \textbf{P4Transformer + Uni4D$^\dagger$}   & \textbf{93.38} \scriptsize (\textcolor{red}{+2.44})\\
    \bottomrule
  \end{tabular} 
\vspace{-0.2cm}
\end{wraptable}

\textbf{MSR-Action3D}. We use the same setting as ~\cite{fan2021point,fan2022point} during fine-tuning on this dataset. The result is reported in Tab. \ref{tab:msr}. Our pre-trained Uni4D significantly improves the baseline trained from scratch, increasing the accuracy rate from $90.94\%$ to $93.38\%$, and performs better than MaST-Pre with $+2.09\%$ in accuracy. Additionally, we evaluate Uni4D under a transfer learning setting by pre-training on NTU-RGBD and then fine-tuning on MSR-Action3D. Interestingly, this model indicates no performance decline. This comparison suggests that our self-disentangled modeling strategy enables the learning of high-quality 4D representations, which are independent of the scale and quality of datasets. The linear probing performance is reported in Tab.~\ref{tab:motion_compare}, where Uni4D significantly outperforms MaST-Pre by $+20.82\%$ accuracy. 

\begin{wraptable}{r}{7cm}
\vspace{-0.2cm}
\centering
\footnotesize
\captionof{table}{Accuracy $(\%)$ on NTU under end-to-end fine-tuning and semi-supervised settings.}
\label{tab:ntu}
\scalebox{0.96}{
\begin{tabular}{lc}
\toprule
    Method & Accuracy \\
    \midrule
    3DV-PointNet++~\cite{wang20203dv} & 84.8 \\
    3DV-Motion++~\cite{wang20203dv} & 84.5 \\
    PST-Transformer~\cite{fan2022point} & 91.0 \\
    kinet~\cite{Zhong_2022_CVPR} & 92.3 \\
    PSTNet~\cite{fan2021pstnet} & 90.5\\
    \rowcolor{gray!10}
    P4Transformer~\cite{fan2021point} & 90.2 \\
    \midrule
    \textit{(End-to-end Fine-tuning)} \\
    \rowcolor{gray!10}
    P4Transformer + MaST-Pre~\cite{shen2023masked}  & 90.8 \\
    \rowcolor{gray!30}
    \textbf{P4Transformer + Uni4D} & \textbf{90.7}\scriptsize (\textcolor{red}{+0.5})\\
    \midrule
    \textit{($50\%$ Semi-supervised)} \\
    \rowcolor{gray!10}
    P4Transformer~\cite{fan2021point} & 81.2 \\
    \rowcolor{gray!30}
    \textbf{P4Transformer + Uni4D} & \textbf{86.5} \scriptsize (\textcolor{red}{+5.3})\\
    \bottomrule
\end{tabular}
}\vspace{-0.2cm}
\end{wraptable}

\textbf{NTU-RGBD}. Due to the large scale of this dataset, we pre-trained our model for only 100 epochs. As listed in Tab. \ref{tab:ntu}, Uni4D improves the baseline for $+0.5\%$ accuracy and obtains comparable performance to MaST-Pre, which has been pre-trained for 200 epochs. This comparison further validates the effectiveness of our approach.

Although modeling inconsistent motion patterns, MaST-pre~\cite{shen2023masked} achieves improvements in action recognition (e.g., $91.29\%$ on MAR-Action3D). This is because this task emphasizes coarse-grained 4D representation within short-term videos~\cite{Liu_2022_CVPR}. We then demonstrate its inability to process long-term videos.

\textbf{HOI4D}. This dataset, with a focus on fine-grained action segmentation, is more challenging than the above two benchmarks. Tab. \ref{tab:hoi4d} reports the segmentation performance of our method following the setting in~\cite{zhang2023complete}. It is evident that Uni4D outperforms all models whether fully-supervised, self-supervised, or fine-tuned, across all metrics. In particular, Uni4D gets $+3.6\%$ accuracy improvements and $+4.8\%$ F1 score at the overlapping threshold of $50\%$ over our baseline. This verifies that our pre-trained encoder has learned discriminative representations and is transferable to comprehensive 4D tasks.

\begin{table}
\centering
\small
\caption{Action segmentation performance on HOI4D. We report the frame-wise accuracy (Acc.), segmental edit distance (Edit), and segmental F1 scores at three levels of threshold.}
\label{tab:hoi4d}
\vspace{0.2cm}
\scalebox{0.95}{
  \begin{tabular}{lcccccc}
    \toprule
    Method & Frames & Acc. & Edit & F1@10 & F1@25 & F1@50 \\
    \midrule
    P4Transformer~\cite{fan2021point} & 150  & 71.2& 73.1 &73.8& 69.2& 58.2\\
    C2P~\cite{zhang2023complete} & 150 & 73.5& 76.8& 77.2& 72.9 &62.4\\
    \rowcolor{gray!10}
    PPTr~\cite{wen2022point} & 150 & 77.4& 80.1 &81.7 
    &78.5 &69.5\\
    \midrule
    PPTr + STRL~\cite{2021Spatio} & 150 & 78.4& 79.1& 81.8 &78.6 &69.7\\
    PPTr + VideoMAE~\cite{feichtenhofer2022masked} & 150 & 78.6&  80.2 &81.9 &78.7 &69.9\\
    \midrule
    \rowcolor{gray!20}
    \textbf{PPTr + Uni4D} & 150 & \textbf{81.0}\scriptsize (\textcolor{red}{+3.6})& \textbf{82.6}\scriptsize (\textcolor{red}{+2.5}) &\textbf{84.6}\scriptsize (\textcolor{red}{+2.9})& \textbf{82.2}\scriptsize (\textcolor{red}{+3.7})& \textbf{74.2}\scriptsize (\textcolor{red}{+4.8})\\
    \bottomrule
  \end{tabular}
  }
\end{table}

\subsection{Semi-supervised Learning}
We then conduct semi-supervised learning experiments on NTU-RGBD. Following ~\cite{shen2023masked}, we use the full dataset during pre-training and then fine-tune with $50\%$ data for 20 epochs. As reported in Tab. \ref{tab:ntu}, the baseline trained from scratch fails to learn expressive 4D representations for recognizing various actions with limited data. In contrast, our pre-trained encoder improves P4Transformer by $+5.3\%$. This clearly demonstrates that Uni4D learns high-quality representations.

\begin{wraptable}{r}{8cm}
\vspace{-3mm}
\centering
\small
\caption{Gesture recognition accuracy $(\%)$ fine-tuned on NvGesture (NvG) and SHREC'17 (SHR). }
\label{tab:gesture}
  \begin{tabular}{lcc}
    \toprule
    Method & NvG & SHR\\
    \midrule
    FlickerNet~\cite{2019FlickerNet} & 86.3 & -\\
    PointLSTM~\cite{min_CVPR2020_PointLSTM}  & 85.9 & 87.6\\
    PointLSTM-PSS~\cite{min_CVPR2020_PointLSTM} & 87.3 & 93.1\\
    kinet~\cite{Zhong_2022_CVPR} & 89.1 & 95.2\\
    \rowcolor{gray!10}
    P4Transformer~\cite{fan2021point} & 87.7& 91.2\\
    \midrule
    \rowcolor{gray!10}
    P4Transformer + MaST-Pre~\cite{shen2023masked} & 89.3 & 92.4 \\
    \rowcolor{gray!20}
    \textbf{P4Transformer + Uni4D} & \textbf{89.6}\scriptsize (\textcolor{red}{+1.9}) & \textbf{93.8}\scriptsize (\textcolor{red}{+2.6})\\ 
    \bottomrule
  \end{tabular}
  \vspace{-0.3mm}
\end{wraptable}
\subsection{Transfer Learning}
We evaluate the generalization ability of the learned representation by pre-training the model on NTU-RGBD, then fine-tuning on NvGesture and SHREC'17. This experiment is not only cross-dataset but also cross-task, as the model is pre-trained on action recognition while fine-tuned to perform gesture recognition. The recognition accuracy is presented in Tab.~\ref{tab:gesture}. Uni4D gets $+1.9\%$ improvements on NvG and $+2.6\%$ on SHREC'17 over the baseline, outperforming another pre-training approach, MaST-Pre. This substantiates Uni4D as a unified framework for 4D tasks, effectively modeling discriminative and transferable representations.

\subsection{Few-shot Learning}
\label{sec:few_show_learning}

\begin{wraptable}{r}{8cm}
\vspace{-4mm}
\centering
\caption{Action recognition accuracy ($\%$) on MSR-Action3D under few-shot learning setting. }
\label{tab:few-shot}
\footnotesize
\scalebox{1}{
  \begin{tabular}{cccccc}
    \toprule
    \multirow{2}{*}{Method} & \multicolumn{2}{c}{\textbf{\textit{5-way}}} & & \multicolumn{2}{c}{\textbf{\textit{10-way}}} \\
    \cline{2-3}\cline{5-6}
    &\textit{1-shot}&\textit{5-shot}&&\textit{1-shot}&\textit{5-shot}\\
    \midrule
     MaST-Pre~\cite{shen2023masked}  & 70.2 & 95.7 && 71.1 & 92.7 \\
     \rowcolor{gray!20}
     \textbf{Uni4D}  & \textbf{74.5} & \textbf{97.8} && \textbf{79.4} &\textbf{95.8} \\
     \textit{Improve} & \textcolor{red}{+4.3} & \textcolor{red}{+2.1} && \textcolor{red}{+8.3} & \textcolor{red}{+3.1} \\
    \bottomrule
  \end{tabular}
  \vspace{-4mm}
}
  \vspace{-0.3mm}
\end{wraptable}
To further demonstrate the superiority of our method, we conduct few-shot learning experiments, adopting the \textit{``n-way m-shot"} setting from \cite{pang2022masked,yu2021pointbert}. To our knowledge, we are the first to conduct a few-shot learning experiment for 4D tasks. For action recognition in point cloud videos, we define \textit{``shot"} as one video, and \textit{``way"} as one specific action. For example, \textit{``5-way 1-shot"} indicates the input of five actions, each action only has one video during training. Tab.~\ref{tab:few-shot} reports the few-shot learning performance of Uni4D compared to MaST-Pre. Each method is pre-trained on NTU, then fine-tuned on MSR under different few-shot learning settings. The proposed method exhibits dominant superiority across all settings. Uni4D obtains absolute improvements over MaST-Pre, indicating better one-shot and few-shot learning ability. Notably, Uni4D surpasses MaST-pre by $+8.3$ recognition accuracy under \textit{``10-way 1-shot"} as the strictest setting. This comparison fully demonstrates that the Uni4D framework is effective in extracting discriminative features.


\section{Ablation Studies}
\label{sec:ab}

\textbf{Architecture Design}. Our approach introduces two learnable tokens to disentangle features within the decoder. To further validate this design, we consider three variants: (1) The baseline model A0 is trained solely on the geometry reconstruction objective; (2) model A1 removes the latent token and relies solely on the geometry token; (3) model A2 uses two separate decoders for geometry reconstruction and semantic preservation, respectively. We compute Chamfer distance to measure reconstruction accuracy, and cosine similarity between $\boldsymbol{Z}_{lat}$ and $\boldsymbol{Z}_{geo}$ to assess disentanglement. 

As listed in Tab.~\ref{tab:ab_architecture}, model A0 achieves the lowest Chamfer distance, consistent with its training objective. Suffering from feature entanglement, model A1 indicates a high cosine similarity and fails to improve fine-tuning performance. Model A2 achieves the lowest cosine similarity, suggesting better disentanglement. Nevertheless, this is primarily attributed to two decoders independently learning geometry and motion representations. While this design effectively achieves disentanglement, it undermines the learning ability of the encoder and introduces additional complexity. In contrast, our full model A3 improves fine-tuning performance from $91.29\%$ to $93.28\%$. This verifies the effectiveness of our self-disentangled learning strategy, where learnable latent tokens facilitate feature disentanglement and add minimal complexity.

\begin{table}
    \centering
    \footnotesize
    \caption{Ablation study on architecture designs. CD: Chamfer Distance. Cos: Cosine similarity.}
    \vspace{1mm}
    \begin{tabular}{l|ccc|ccc}
    \toprule
     & Geometry Token & Latent Token & Shared Decoder & CD $\times10^{-3}$ ($\downarrow$) & Cos. ($\downarrow$) & Acc. ($\uparrow$) \\
    \midrule
    \rowcolor{gray!10}
    A0 &  &  &  & \textbf{2.2} & - & 91.63 \\
    A1 & \ding{52} & \ding{56} & \ding{52} & 5.7 & 0.9991 & 90.24 \\
    A2 & \ding{52} & \ding{52} & \ding{56} & 3.9 & \textbf{0.0301} & 90.94 \\
    A3 & \ding{52}& \ding{52} & \ding{52} & 4.3 & 0.2536 & \textbf{93.38} \\
    \bottomrule
    \end{tabular}
    \label{tab:ab_architecture}
    \vspace{-4mm}
\end{table}

\begin{wraptable}{r}{7.7cm}
\vspace{-0.3cm}
    \small
    \centering
  \caption{Ablation studies on different pretext tasks.}
  \label{tab:ab_pretext}
  \scalebox{0.9}{
  \begin{tabular}{cccccc}
    \toprule
    Model&$\mathcal{L}_{geo}$&$\mathcal{L}_{lat}$&$\mathcal{L}_{global}$&$\mathcal{L}_{motion}$&Acc.(\%)\\
    \midrule
    B1&  \ding{52} & &  & &91.63\\
    B2& \ding{52}& \ding{52}& $ $ & $ $&85.36\\
    B3& \ding{52}& $ $& $ $ & \ding{52}&91.63\\
    B4& \ding{52}& \ding{52}& $ $ & \ding{52}&90.94\\
    B5& \ding{52}& \ding{52}& \ding{52} & $ $&91.29\\
    B6& $ $& \ding{52}& \ding{52} & \ding{52}&91.63\\
    B7 (Ours)& \ding{52}& \ding{52}& \ding{52} & \ding{52}&\textbf{93.38}\\
  \bottomrule
\end{tabular}
}
  \vspace{-0.1cm}
\end{wraptable}

\textbf{Pretext task}.
We construct ablation studies of different pretext tasks on MSR-Action3D. As listed in Tab.~\ref{tab:ab_pretext}, the baseline model B1 performs only geometric reconstruction. Notably, model B2, which incorporates only the latent alignment objective, exhibits a significant performance drop. This result highlights the inherent gap between motion and geometry: the decoder struggles to predict low-level structural details while preserving high-level semantics. Meanwhile, model B3 introduces motion alignment in the latent space but fails to improve fine-tuning performance. We attribute this to the overfitting issue during decoding. The pre-trained encoder thereby models suboptimal 4D representations, which affects the fine-tuning effectiveness for downstream tasks like action recognition.

\begin{wrapfigure}{r}{0.55\textwidth}
\vspace{-5mm}
\begin{center}
\includegraphics[clip,trim=0 0 0 0,width=0.55\textwidth]{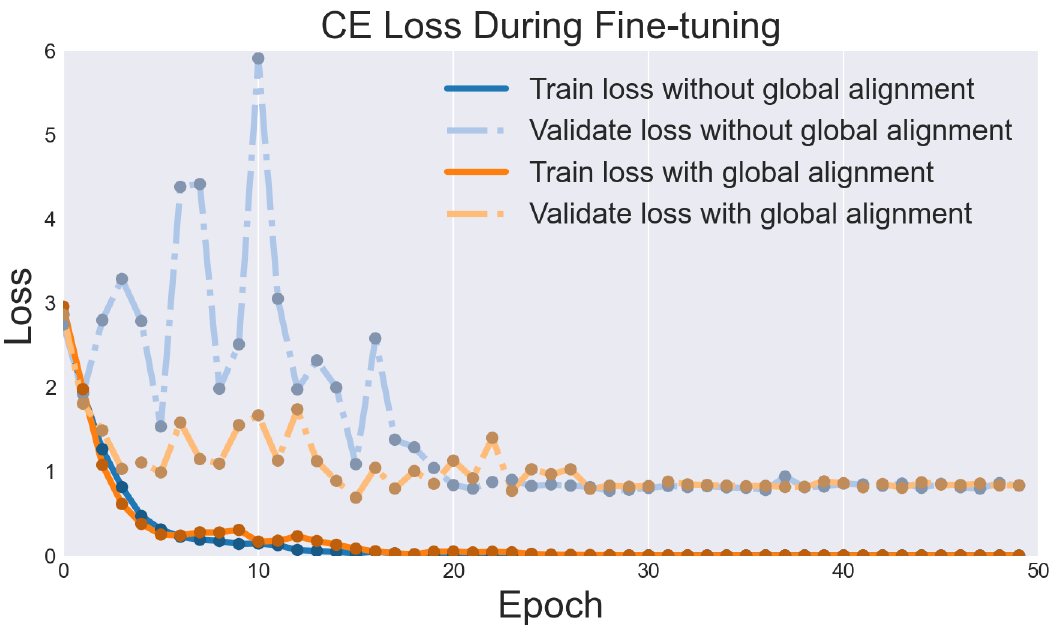}
\end{center}
\vspace{-3mm}
\caption{Ablation study of global alignment. We report the train and validate losses during fine-tuning.}
\label{fig:CELOSS}
\end{wrapfigure}

Interestingly, when adding a single alignment objective in the latent space, model B4 suffers a larger performance drop than model B5. To understand what affects recognition accuracy, we ablate the global alignment module and plot the losses during fine-tuning on MSR-Action3D in Fig.~\ref{fig:CELOSS}. Without global alignment, the pre-trained model lacks the ability to capture video-level semantics required for action recognition. Consequently, it requires more epochs to regain this capacity, as evidenced by an extremely turbulent validate loss. 

Another observation from Tab.~\ref{tab:ab_pretext} is that model B6, which combines all high-level objectives but removes the geometric reconstruction loss, fails to improve fine-tuning performance. This highlights the indispensable role of low-level geometry in 4D downstream tasks. Model B7, our full model integrating geometric reconstruction, latent alignment, and both motion and global alignment objectives, achieves the highest performance. These results validate the effectiveness of each module of our approach. The above observations also lead to our central argument: while geometry and motion are fundamental for point cloud video understanding, effective 4D representation must go beyond them to capture richer, higher-level, and comprehensive semantics. 

In Appendix~\ref{appendix:ablation}, we provide additional ablation experiments and more discussions.

\section{Visualizations}

\begin{figure}[t]
    \centering
    {\includegraphics[width=1\textwidth]{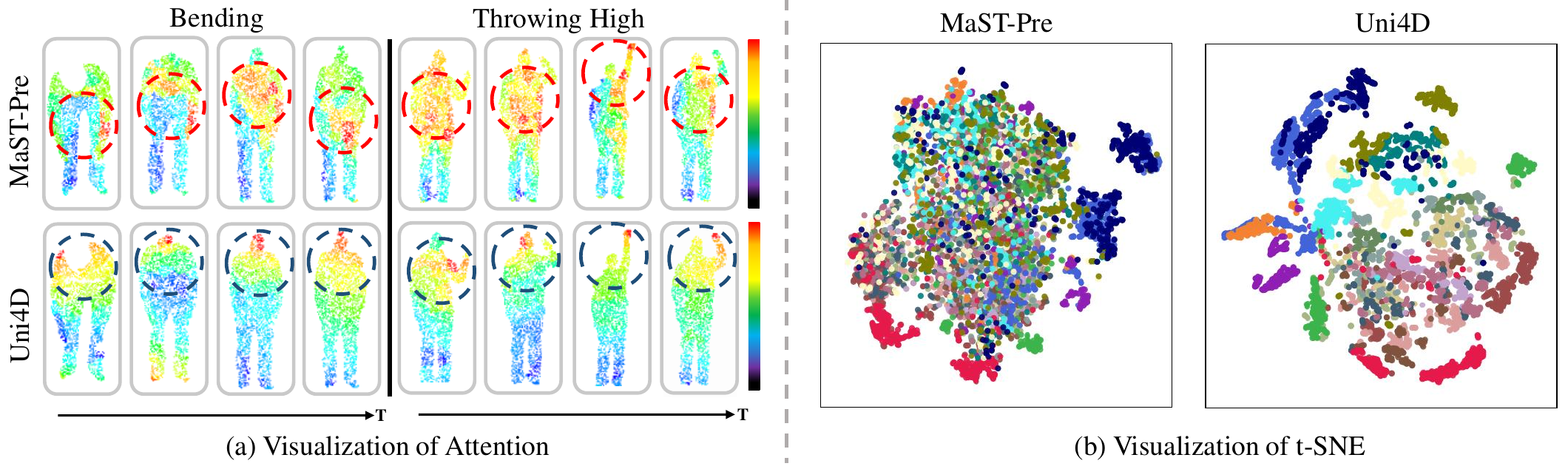}}
	\caption{We compare the learned representation of Uni4D and MaST-Pre through visualization of attention and t-SNE. Note features are obtained from the pre-trained encoder \textbf{without fine-tuning}.}
	\label{fig:VisualizationCompare} 
\vspace{-0.2cm}
\end{figure}

\begin{wrapfigure}{r}{0.5\textwidth}
\vspace{-7mm}
\begin{center}
\includegraphics[clip,trim=0 0 0 0,width=0.48\textwidth]{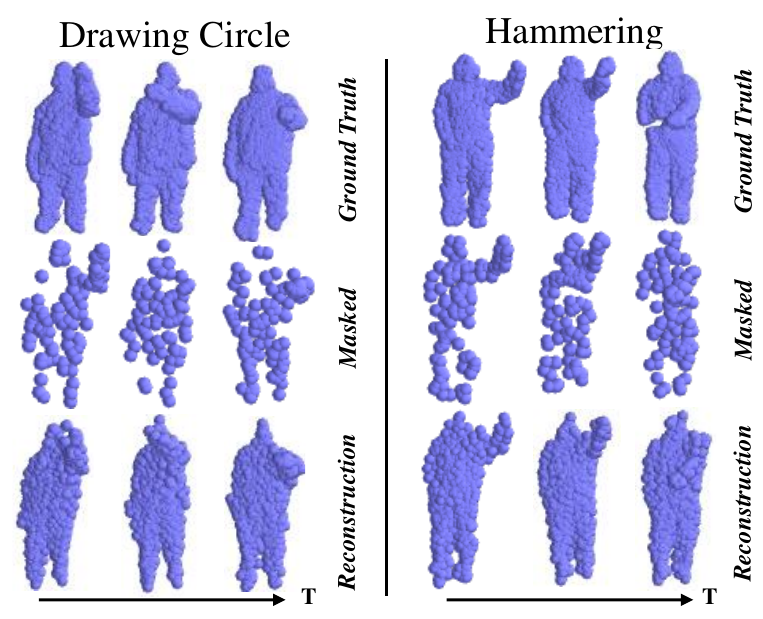}
\end{center}
\vspace{-2mm}
\caption{Visualization of reconstruction results.}
\label{fig:reconstruction}
\end{wrapfigure}

In Fig.~\ref{fig:VisualizationCompare}(a), we visualize the attention activations extracted from the pre-trained encoder. Compared to MaST-Pre, Uni4D shows consistent motion patterns across consecutive frames, indicating long-term temporal dependencies. Another observation is that Uni4D consistently highlights the hand area for the action \textit{``throwing high"}, which suggests our pre-trained encoder produces expressive features to recognize the action. To verify that Uni4D learns meaningful 4D representations during pre-training, we use t-SNE and visualize the feature distributions of the pre-trained encoder. Compared to MaST-Pre, it is evident in Fig.~\ref{fig:VisualizationCompare}(b) that our pre-trained encoder exhibits clear boundaries between different categories, which substantiates the learning of discriminative features.

Fig.~\ref{fig:reconstruction} presents two reconstruction examples of Uni4D pre-trained on MSR-Action3D. Observe that the results have undesired geometric noise. Due to the representation gap, richer semantics may not be conducive to reconstruction accuracy. Appendix ~\ref{appendix:discussion} provides an in-depth discussion.

\section{Limitation and Discussion}
\label{limitation}
Our approach employs two encoders to model expressive semantics, which introduces additional complexity to the pre-training stage, as reported in Tab.~\ref{tab:motion_compare}. This is particularly significant when scaling to large-scale datasets such as~\cite{shahroudy2016ntu}. Nevertheless, our self-disentangled learning strategy accelerates the model in producing meaningful spatio-temporal representations, achieving comparable performance to MaST-Pre~\cite{shen2023masked} even with only half of the training epochs. In future work, we will seek solutions to improve pre-training efficiency to handle large-scale scenarios.

\section{Conclusion}
In this paper, we introduced Uni4D, the first self-disentangled MAE for point cloud video representation learning, addressing key limitations in 4D modeling with two core designs. First, our approach aligns high-level semantics in the latent space with high-level alignment objectives, and reconstructs geometry in Euclidean space. Second, we introduce the latent tokens along with the geometric tokens in a shared decoder to disentangle high-level and low-level features. Extensive experiments validate the effectiveness of Uni4D in learning expressive spatio-temporal 4D representations. Notably, our pre-trained encoder models long-term temporal dependencies, which are essential for fine-grained 4D downstream tasks. We believe our self-disentangled MAE offers valuable insights to a wide range of tasks, such as video understanding and multimodal learning.

\bibliographystyle{unsrt}  
\small
\bibliography{Reference}
\normalsize

\appendix
\clearpage
\setcounter{page}{1}

\appendix

\section{Network Details}
\label{appendix:architecture}
\subsection{Point 4D Convolution}
\label{theory}
Following~\cite{fan2021point,shen2023masked}, we use point 4D convolution as our embedding network. Let $P_t \in \mathbb{R}^{N\times 3}$ and $F_t \in \mathbb{R}^{N\times C}$ represent the point coordinates and features of the $t$-th frame in a sequence of input point clouds, where $N$ and $C$ represent the number of points and feature channels, respectively. Thus, the sequence of point clouds can be expressed as $P= \{ P_t,F_t \}_{t=1}^{L}$, where $L$ is the sequence length of point clouds. We employ a 4D Convolution \cite{fan2021point} to capture point features as $P'= \{ P_{t}',F_{t}'\}_{t=1}^{L'}$, with a down-sampled sequence length $L'\leq L$ and down-sampled point number $N'<N$. The above process can be mathematically formulated as:

\begin{equation}
\label{eq:p4dconv}
\small
f_{t}'^{(x,y,z)} = \sum _{t=-r_t}^{r_t} \sum _{\|\delta_x,\delta _y,\delta _z \|< r_s} \left\{(W_d\cdot(\delta_x,\delta_y,\delta_z,\delta_t)^{T})+(W_f\cdot f_{t+\delta_{t}}^{(x+\delta _{x},y+\delta _{y},z+\delta _{z})})\right\},
\end{equation}

where $(x,y,z) \in \mathbb{R}^{N' \times 3}$ is the coordinates of the point cloud, $(\delta _x,\delta _y,\delta _z,\delta _t)$ is the spatial-temporal displacement, $W_d \in \mathbb{R}^{C' \times 4}$ is a transformation matrix, and $\cdot$ denotes matrix multiplication. $f_{t+\delta_{t}}^{(x+\delta _{x},y+\delta _{y},z+\delta _{z})}$ denotes the features of points at $(x +\delta _x ,y +\delta _y ,z + \delta _z, t + \delta_{t})$ and
$W_f \in \mathbb{R}^{C' \times C}$ is to transform spatial-temporal features. $r_s$ and $r_t$ denote the spatial radius and temporal radius, respectively.

Some datasets, such as MSR-Action3D and NTU-RGBD (+60), only provide coordinate information of point clouds. In these cases, Eq.~\ref{eq:p4dconv} can be simplified to:
\begin{equation} \label{2}
f_{t}'^{(x,y,z)} = \sum _{t=-r_t}^{r_t} \sum _{\|\delta _x,\delta _y,\delta _z \|< r_s} \left(W_d\cdot(\delta_x,\delta_y,\delta_z,\delta_t)^{T}\right..
\end{equation}

After point 4D convolution, the input of the encoder is $\boldsymbol{E} =f_{t}'^{(x,y,z)}$, where $x,y,z$ denotes the coordinates of $\hat{p}_i$ in $\boldsymbol{Tube}_{\hat{p}_i}$. For masked reconstruction, the visible embedding is $\boldsymbol{E_v} =f_{t}'^{(x_v,y_v,z_v)}$, where $x_v,y_v,z_v$ denotes the coordinates of visible point clouds.

\subsection{Encoder}
Following~\cite{shen2023masked,han2024masked}, we utilize P4Transformer \cite{fan2021point} as both online encoder and momentum encoder to learn meaningful 4D representations during the pre-training phase. These two encoders are fed with visible embeddings $\boldsymbol{E}_v$ and the unmasked full embeddings $\boldsymbol{E}$, respectively. The encoder is implemented as a spatio-temporal Transformer, which consists of Multi-head Self-Attention (MSA), Layer Normalization (LN), and a MLP block in each layer $l$ as: 
\begin{equation}
f^{l}= MSA(LN(z^{l-1}))+z^{l-1},
\end{equation}
\begin{equation}
z^{l}= MLP(LN(f^{l}))+f^{l}.
\end{equation}
\subsection{Decoder}
Typically, the decoder of MAE is only used for pre-training and will be discarded during downstream fine-tuning. Therefore, it is crucial to adopt a lightweight decoder to maximize the learning ability of the encoder. Instead of using two decoders~\cite{huang2023contrastive,han2024masked}, the proposed self-disentangled MAE adopts the vanilla Transformer as our decoder, which consists of Multi-head Self-Attention (MSA), Layer Normalization (LN), and a MLP block in each layer $l$. To reconstruct geometry, the encoded visible embeddings $\boldsymbol{Z}_v$ are concatenated with the geometry tokens $T_{geo}$, which are learnable during pre-training, denoted $z_{geo}^{0}=[Z_v,T_{geo}]$. To prevent feature entanglement, we innovatively introduce the latent tokens $T_{lat}$. Similarly, the latent tokens are concatenated to the encoded visible embeddings $\boldsymbol{Z}_v$, yielding $z_{lat}^{0}=[Z_v,T_{lat}]$. The operation of each layer $l$ can be expressed as: 
\begin{equation}
f_d^{l}= MSA(LN(z_d^{l-1}))+z_d^{l-1},
\end{equation}
\begin{equation}
z_d^{l}= MLP(LN(f_d^{l}))+f_d^{l}.
\end{equation}
where $z_d^{l}=z_{geo}^{l}$ for geometry reconstruction and $z_d^{l}=z_{lat}^{l}$ for latent alignment. 

After decoding, the geometry tokens are transformed into features $\boldsymbol{Z}_{geo}$ for reconstruction (Eq.~\ref{eq:geometry_loss}), and the latent tokens are transformed into features $\boldsymbol{Z}_{lat}$ for alignment in (Eq.~\ref{eq:latent_loss}).

\section{Datasets}
\label{sec:dataset}
In this section, we briefly introduce five datasets that we have used for evaluation.

\textbf{MSR-Action3D}~\cite{2010Action} serves as a widely accepted benchmark dataset in the domain of action recognition, especially concerning human action recognition from 3D skeletal data. Introduced by Microsoft Research (MSR), this dataset comprises depth sequences, RGB videos, and 3D skeletal data capturing diverse human actions performed by multiple subjects. The MSR-Action3D dataset encompasses 20 actions, a total of 23K frames. We adopt the same training and test set partitioning as previous works~\cite{liu2019meteornet,fan2021point}.

\textbf{NTU-RGBD}~\cite{shahroudy2016ntu} stands as the largest dataset for action recognition, comprising 56K videos encompassing 60 action categories and totaling 4M frames. These videos were captured using Kinect v2, featuring 3 cameras and involving 40 subjects as performers. Following previous work~\cite{fan2021pstnet}, we convert the depth sequences into point cloud sequences to enable the exploration of 3D action recognition tasks in point cloud form. Under the cross-subject setting ~\cite{shahroudy2016ntu}, 40,320 training videos and 16,560 test videos are used. 

\textbf{HOI4D}~\cite{Liu_2022_CVPR} is a large scale 4D self-centered view dataset with rich annotations to facilitate the study of class-level Human-Object Interaction (HOI). HOI4D contains 2.4 million RGB-D egocentric video frames, over 4,000 sequences, collected by four participants interacting with 610 different indoor rooms across 16 categories and 800 different target instances.

\textbf{NvGestures}~\cite{molchanov2016online} is a specialized collection tailored for gesture recognition within driving scenarios, encompassing a diverse array of gestures commonly utilized by drivers to interact with in-car systems while on the road. It comprises 1532 videos, with 1050 allocated for training and 482 for testing, composed of 25 classes.

\textbf{SHREC'17}~\cite{de2017shrec} is comprised of 2800 videos in 28 gestures. This dataset is split into 1960 training videos and 840 test videos.
 
\section{Experiment Details}
\label{appendix:details}
We provide more detailed experimental settings and report configurations in Table~\ref{tab:parameters_all}.
\subsection{Pre-training Setups}
\textbf{MSR-Action3D and NTU-RGBD}. For division and embedding, the temporal down-sampling rate is set to $2$ and the temporal kernel size $l$ of each point tube is set to 3. The spatial down-sampling rate is set to 32. The radius of each support domain $r$ is set to 0.1/0.3 on NTU-RGBD/MSR-Action3D and the number of neighbor points $n$ within the spherical query is set to 32. 

\textbf{HOI4D}.
For division and embedding, the temporal down-sampling rate is set to $1$ and the temporal kernel size $l$ of each point tube is set to 3. The radius of each support domain $r$ is set to 0.9 and the number of neighbor points $n$ within the spherical query is set to 32. 

\textbf{Note}: we remove the global alignment objective during pre-training on HOI4D for action segmentation. This task focuses on fine-grained details where video-level information could harm segmentation accuracy. Therefore, we employ only the bidirectional motion alignment objective to capture frame-level semantics, which performs better than using both alignment objectives.

\subsection{Downstream Fine-tuning Setups}

\textbf{End-to-end fine-tuning on MSR-Action3D}. We adopt the same setting as previous works~\cite{shen2023masked,han2024masked}. The input is a point cloud video consisting of $24$ frames with $2048$ points per frame. The space radius $r_s$ is set to $0.7$ and the temporal radius $r_t$ is set to $3$. We sample $64$ center points, $12$ center frames, respectively. The fine-tuning process consists of $50$ epochs with a batchsize of $48$, and our optimizer selects AdamW with a learning rate of $0.0005$, using a cosine decay strategy where the learning rate is linearly warmed up by $10$ epochs. 

\textbf{End-to-end fine-tuning on NTU}. The input is a point cloud video consisting of $24$ frames with $2048$ points per frame. The space radius $r_s$ is set to $0.1$ and the temporal radius $r_t$ is set to $3$. 
We sample $64$ center points, $12$ center frames, respectively. The fine-tuning process consists of $20$ epochs with a batchsize of $48$, and our optimizer selects AdamW with a learning rate of $0.0005$, using a cosine decay strategy where the learning rate is linearly warmed up by $10$ epochs. 

\textbf{End-to-end fine-tuning on HOI4D}. The input is a point cloud video consisting of $150$ frames with $2048$ points per frame. The space radius $r_s$ is set to $0.9$ and the temporal radius $r_t$ is set to $3$. The fine-tuning process consists of $50$ epochs with a batchsize of $8$, and our optimizer selects SGD with a learning rate of $0.05$, using a decay strategy where the learning rate is linearly warmed up by $5$ epochs. 

\textbf{Semi-supervised learning on NTU}. After pre-training on the full dataset, we perform semi-supervised fine-tuning on half of the data. The fine-tuning process consists of $20$ epochs with a batch size of $48$, and our optimizer selects AdamW with a learning rate of $0.0005$, using a cosine decay strategy where the learning rate is linearly warmed up by $10$ epochs. 

\begin{table}[t]
\centering
\caption{Detailed training configurations and hyperparameters across datasets. ``pre/ft" indicates pre-training / fine-tuning stages.}
\label{tab:parameters_all}
\scalebox{0.9}{
\begin{tabular}{l|ccccc}
\toprule
\textbf{Configuration} & \textbf{MSR} & \textbf{NTU} & \textbf{HOI4D} & \textbf{NvGestures} & \textbf{SHREC'17} \\
\midrule
Pre-train optimizer         & AdamW   & AdamW   & AdamW   & --     & --     \\
Fine-tune optimizer         & AdamW   & AdamW   & SGD     & AdamW  & AdamW  \\
Pre-train learning rate     & 3e-4    & 3e-4    & 5e-4    & --     & --     \\
Fine-tune learning rate     & 5e-4    & 5e-4    & 5e-2    & 5e-4   & 1e-3   \\
Pre-train weight decay      & 5e-2    & 5e-2    & 1e-4    & --     & --     \\
Fine-tune weight decay      & 1e-4    & 5e-2    & 1e-4    & 1e-4   & 1e-4   \\
LR scheduler                & cosine  & cosine  & cosine  & cosine & cosine \\
Epochs (pre/ft)             & 200 / 50& 100 / 20& 50 / 50 & -- / 50& -- / 50\\
Batch size (pre/ft)         & 96 / 48 & 56 / 48 & 4 / 8   & -- / 24& -- / 24\\
Spatial radius $r_s$ (pre/ft) & 0.3 / 0.7 & 0.1 / 0.1 & 0.9 / 0.9 & -- / 0.1 & -- / 0.3 \\
Temporal stride $r_t$       & 2       & 2       & 1       & 2      & 2      \\
Temporal radius             & 3       & 3       & 3       & 3      & 3      \\
Neighbor samples            & 32      & 32      & 32      & 16     & 9      \\
Encoder heads / depth       & 8 / 5   & 8 / 10  & 8 / 5   & 8 / 10 & 8 / 10 \\
Momentum parameter          & 0.999   & 0.999   & 0.999   & --     & --     \\
Queue size                  & 12288   & 14436   & --      & --     & --     \\
Decoder heads / depth       & 8 / 4   & 8 / 4   & 8 / 4   & --     & --     \\
Temperature $\tau$          & 0.1     & 0.1     & 0.1     & --     & --     \\
Few-shot / Semi-sup. LR     & 5e-4    & 5e-4    & --      & --     & --     \\
Input points (pre/ft)       & 1024 / 2048 & 1024 / 2048 & 2048 / 2048 & -- / 512 & -- / 256 \\
\bottomrule
\end{tabular}
}
\end{table}

\textbf{Few-shot learning on MSR-Action3D}. As mentioned in the main paper, we innotatively introduce a few-shot learning experiment for 4D tasks to validate the effectiveness of our method. Tailored to 4D action recognition, we define the typical ``K-way N-shot" settings as ``shot" for video and ``way" for action. Specifically, MSR-Action3D dataset contains 12 videos in each action class, i.e., ``1-way 12-shot". Each model is pre-trained on NTU and fine-tuning on MSR-Action3D with a few-shot learning setting. We choose 5-way and 10-way to randomly take 1-shot and 5-shot, respectively. The fine-tuning process consists of $50$ epochs with a batchsize of $48$, and our optimizer selects AdamW with a learning rate of $0.0005$, using a cosine decay strategy where the learning rate is linearly warmed up by $10$ epochs. 

\textbf{Transfer learning on NvGestures and SHREC'17}
During fine-tuning, an AdamW optimizer with a batch size of 24 is used. The initial learning rate is set to 0.0005/0.001 with a cosine decay strategy on NvGestures/SHREC'17, respectively. The pre-trained model is fine-tuned on NvGesture and SHREC'17 for 50 epochs.

\section{Additional Ablation Studies}
\label{appendix:ablation}

We conduct additional ablation studies on the masking strategy and different hyperparameters below. 

\textbf{Masking Ratio \& Strategy}
We begin by studying the impact of masking ratios and masking strategies on model performance. As shown in Table~\ref{tab:appendix_hyperparameter_ablation}, we experiment with three different masking ratios (65\%, 75\%, 85\%). Due to the sparse nature of point clouds, the encoder fails to learn good 4D representations from limited unmasked embeddings, with poor performance under a high masking ratio (85\%). On the other hand, if the masking ratio is low (65\%), the encoder may not learn meaningful spatio-temporal information. We also conduct experiment with various masking strategy, including block masking strategy, video-level masking strategy, and frame-level masking strategy. The video-level masking strategy applies a global mask across the entire video, resulting in potentially varied masking ratios across frames. Frame-level masking strategy means each frame will be masked at the same ratio. Block masking strategy indicates that adjacent areas will be masked in frame. The masking ratio is set to 75\% when comparing different masking strategies. Unlike MaST-Pre \cite{shen2023masked} that employs a video-level masking strategy, we utilize a frame-level masking strategy as our default setting during comparison. 

\begin{wraptable}{r}{7cm}
\vspace{-0.4cm}
\centering
\caption{Ablation study on MSR-Action3D. Accuracy (\%) under different hyperparameter and design settings.}
\label{tab:appendix_hyperparameter_ablation}
\small
\scalebox{0.9}{
\begin{tabular}{lcc}
\toprule
\textbf{Setting} & \textbf{Configuration} & \textbf{Accuracy (\%)} \\
\midrule
\multirow{3}{*}{Masking Ratio} 
  & 65\% & 91.99 \\
  & \textbf{75\%} & \textbf{93.38} \\
  & 85\% & 89.20 \\
\midrule
\multirow{3}{*}{Masking Strategy} 
  & block & 88.94 \\
  & video & 92.33 \\
  & \textbf{frame} & \textbf{93.38} \\
\midrule
\multirow{3}{*}{Temperature $\tau$} 
  & $[0.01,0.05)$ & 90.24 \\
  & $[0.05,0.1)$ & 91.98 \\
  & $\mathbf{0.1}$ & \textbf{93.38} \\
\midrule
\multirow{3}{*}{Momentum $m$} 
  & $[0.99,0.995)$ & 85.36 \\
  & $[0.995,0.999)$ & 91.29 \\
  & $\mathbf{0.999}$ & \textbf{93.38} \\
\midrule
\multirow{3}{*}{Queue Size} 
  & 6144 & 90.94 \\
  & \textbf{12288} & \textbf{93.38} \\
  & 24576 & -- \\
\bottomrule
\end{tabular}
}
\end{wraptable}

\textbf{Temperature (\(\tau\))}:
The temperature parameter \(\tau\)) is critical for bidirectional motion alignment and global alignment, which affect the learning ability of encoder to distinguish between positive and negative pairs. We consider three different settings: temperature is set to $[0.01,0.05)$, $[0.05,0.1)$ and $0.1$. We found that when temperature is set to \(0.1\) yields the best performance. This setting effectively balances the contrastive loss function, ensuring appropriate differentiation between positive and negative samples. 

\textbf{Momentum Parameter}: The momentum parameter is a key component in two-tower frameworks. It controls the update rate of the momentum encoder by performing an exponential moving average (MEA) of the parameters from the online encoder. The optimal setting of \(0.999\) achieves the best performance. This ensures a stable and adaptable update for the momentum encoder. Other values may lead to excessive fluctuations in latent representation due to rapid updates, reducing the encoder’s ability to learn meaningful latent representation.

\textbf{Queue Size ($K$)}: We design a dynamic queue to store a large set of negative samples for learning discriminative representation at video-level. A size of \(12,288\) achieves the highest accuracy, suggesting that an adequately large queue enhances contrastive learning by maintaining a sufficient pool of negative samples. A smaller queue (\(6,144\)) results in insufficient negative samples, limiting the model's capacity to learn robust discriminative representations. Conversely, an excessively large queue (\(24,576\)) may exceed the limits of device memory.

\section{Additional Discussion}
\subsection{Discussion on Reconstruction \& Alignment}
\label{appendix:discussion}
Most existing self-supervised learning methods for static point clouds adopt geometric reconstruction as a pretext task, and have demonstrated highly competitive performance across various downstream tasks~\cite{pang2022masked, PointContrast2020}. However, dynamic point clouds exhibit significantly higher spatio-temporal complexity, making pure geometry-based reconstruction insufficient to capture comprehensive semantics in 4D data. Some recent efforts attempt to address this limitation by incorporating external knowledge for motion modeling in the design of pretext tasks, such as temporal cardinality difference and motion trajectory~\cite{shen2023masked,han2024masked}. In contrast, we propose a novel paradigm that aligns video-level and frame-level features in the latent space, aiming to enhance the model's understanding of spatio-temporal representations of point cloud videos. Unlike previous methods that pursue accurate reconstruction in Euclidean space, our approach, with a focus on temporal consistency in the latent space, inevitably introduces a degradation of reconstruction quality.

\begin{figure}
\vspace{-5mm}
	\centering
        \label{MSRvisualization}{
		\includegraphics[width =0.8\textwidth]{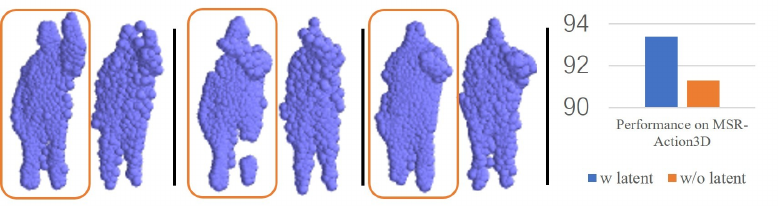} }\\
	\caption{Reconstruction comparison between models trained only with reconstruction objective (marked in orange) and combined with alignment objectives.}
	\label{ReconstrutionComparsion} 
    \vspace{-4mm}
\end{figure}  

To provide in-depth insights, we provide reconstruction examples of models trained with or without high-level alignment objectives in Fig~\ref{ReconstrutionComparsion}. By removing high-level alignment objectives, the model performs better on reconstruction but has lower accuracy on MSR-Action3D ($\sim91\%$). In fact, our full model produces noisy reconstructions but has demonstrated superiority under fine-tuning ($\sim93\%$). This finding provides valuable insights for understanding 4D representations and developing effective SSL approaches for point cloud videos: \textbf{the ultimate goal of pre-training is not to focus on better reconstruction accuracy but to capture meaningful spatial-temporal information for better fine-tuning performance}. Nevertheless, low-level reconstruction still plays a critical role during fine-tuning, as the geometric structure is also essential for downstream tasks. This is evident in the comparison of models B6 and B7 in Tab.~\ref{tab:ab_pretext}.

\begin{wrapfigure}{r}{0.35\textwidth}
\vspace{-5mm}
        \label{MSRvisualization}{
		\includegraphics[width =0.35\textwidth]{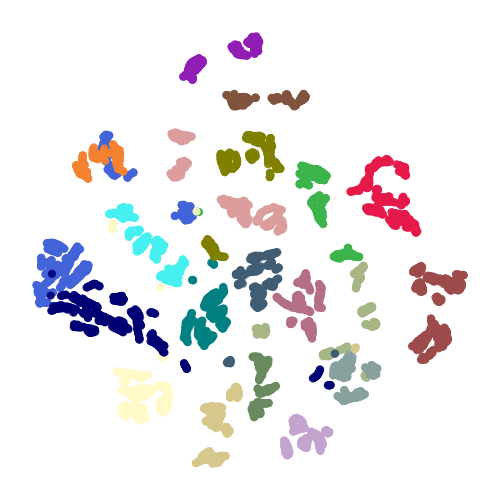} }
    \vspace{-6mm}
	\caption{t-SNE visualization after fine-tuning on MSR-Action3D.}
	\label{FT} 
    \vspace{-15mm}
\end{wrapfigure}

We further visualize the t-SNE features after fine-tuning. As shown in Fig.~\ref{FT}, our method demonstrates superior performance with well-clustered and discriminative feature distributions.

\subsection{Discussion on Limitation}
\label{more limitations}
Our approach employs a momentum encoder along with the online encoder to capture expressive semantics, which inevitably leads to increased memory consumption. We report the pre-training efficiency of Uni4D in Tab.~\ref{tab:appendix_memory}. This may pose challenges for deployment on older devices or memory-constrained environments. In future work, we aim to optimize the model architecture to improve pre-training efficiency.

\begin{table}
\centering
\small
\caption{Comparison of pre-training efficiency on MSR-Action.}
\label{tab:appendix_memory}
  \begin{tabular}{cccc}
    \toprule
     Method &Time (1 epoch)& Memory (MB) & Accuracy (\%)\\
    \midrule
    MaST-Pre~\cite{shen2023masked} & \textbf{17}s & \textbf{15059} &90.94\\
    Uni4D & 26s & 36044 &\textbf{93.38}\\
    \bottomrule
  \end{tabular}
\end{table}

\section{Broader Impact}
\label{Broader Impact}
Uni4D demonstrates a significant improvement over existing methods and establishes new state-of-the-art performance across a variety of tasks. Our self-supervised framework, learning 4D representations without relying on hand-crafted designs or labeled data, is especially useful for applications where annotations are limited, such as robotics, autonomous driving, and healthcare. The learning of long-term temporal dependencies in point cloud videos is valuable in advancing technologies that rely on precise recognition and segmentation. However, like many other vision models, it might include potential negative social impacts. While our method does not use identity-related data, we encourage responsible use and consideration of privacy and fairness in real-world deployments or other sensitive applications. We believe that Uni4D offers a strong step toward a more general understanding of 4D representation, with potential benefits across both academia and industry.



\end{document}